\documentclass{article}

\usepackage[dblblindworkshop, final,nonatbib]{neurips_2026}

\usepackage[utf8]{inputenc} % allow utf-8 input
\usepackage[T1]{fontenc}    % use 8-bit T1 fonts
\usepackage{hyperref}       % hyperlinks
\usepackage{url}            % simple URL typesetting
\usepackage{booktabs}       % professional-quality tables
\usepackage{amsfonts}       % blackboard math symbols
\usepackage{nicefrac}       % compact symbols for 1/2, etc.
\usepackage{microtype}      % microtypography
\usepackage{xcolor}         % colors
\usepackage{biblatex}
\title{Towards Model as a Library: Offline, Community-Sourced AI for Low-Resource African Languages}

\author{%
  Fendji K. E. Jean Louis \\
  Centre of Research, Experimentation and Production\\
  SCEMI, University of Ngaoundere, Ngaoundere, Cameroon \\
  Stellenbosch Institute for Advanced Study \\
  Wallenberg Research Centre at Stellenbosch University, Stellenbosch, South Africa\\
  \texttt{jl.fendji@egcim-univ-ndere.cm} \\
}

\workshoptitle{GlobalSouthAI}

\begin{document}

\maketitle

\begin{abstract}
Large language models are frequently proposed as a route to AI-powered services for African communities, but they are least reliable exactly where the need is greatest: all African languages remain low-resource by any standard measure, and models trained on scraped, standardised text systematically misrepresent the dialectal and regional variation of how people actually speak. We introduce \textbf{Model as a Library (MaaL)}, a software architecture that packages small, community-enrolled speech models as versioned on-device dependencies, enabling offline structured data collection that cannot generatively hallucinate, for populations that current language models serve worst. Rather than relying on web-scraped corpora, MaaL's vocabulary is enrolled directly from a small number of example recordings by the speakers themselves, at the point of deployment. We describe the architecture and its central mechanism -- keyword spotting that turns a closed-vocabulary text form into a voice form, filled and submitted entirely on-device -- and propose transpiling the closed-vocabulary elements already present in widely-deployed digital form tools into MaaL schemas, a low-friction path to voice-first, offline data collection for the low-literacy populations these tools already reach. This is a position and system-design paper: we describe the concept, the mechanism, and an analytical feasibility case, and identify what a working implementation still requires.
\end{abstract}

\textbf{Keywords:} offline AI, keyword spotting, low-resource languages, on-device inference, few-shot learning, human-computer interaction

\section{Introduction}

Large language models are increasingly proposed as infrastructure for education, health, and agricultural services across the Global South. Yet the evidence on their reliability for African languages is not encouraging. African languages remain low-resource by any standard measure, and lower resource levels directly translate into lower-quality, less reliable models -- an inversion in which the populations with the most to gain are served worst. The failure mode is not abstract: on the WARRI benchmark for West African Pidgin, models adapted to the standardised media register (BBC variety) score 76.3--83.4 ChrF++, but the same models drop to approximately 54 on everyday community-level Naija -- a 24.3-point gap that causes model reasoning to drift from accuracy toward linguistic guesswork whenever input departs from the training distribution~\cite{adelani2025warri, clinical_ai_fragility}. This "Standard English bias" is driven by training data that overrepresents one register and erases the rest. Machine-translation pipelines, often proposed as a workaround, introduce their own compounding and poorly understood errors. The underlying cause is not a prompting problem; it is data scarcity, and no amount of scale fixes a corpus that was never representative to begin with.

This is a structural, not incidental, problem for deployment. Large parts of the population most in need of these tools also face the least reliable connectivity and linguistic coverage simultaneously -- so even where a capable multilingual model exists, connectivity and language together block access, widening the very data gap that caused the problem~\cite{fendji2024leftbehind}. A tool requiring a live connection to a distant server is, for this population, not a language-access solution but a second barrier stacked on the first.

We argue that for a large and important class of applications -- structured data collection, not open-ended dialogue -- neither of these failure modes is necessary. We introduce \textbf{Model as a Library (MaaL)}, a system architecture in which trained models are packaged as versioned, on-device software dependencies, analogous to code libraries, rather than accessed as a remote service. A MaaL \emph{domain model} is a small keyword-spotting classifier whose vocabulary is enrolled directly from a small number of example recordings -- typically 5--10 in published few-shot keyword-spotting work~\cite{parnami2022few,rusci2023fewshot} -- by the speakers themselves, collected on-device, with no cloud round-trip and no retraining. Because inference is closed-set matching rather than open-ended generation, MaaL cannot hallucinate a response in the way a generative model can: it returns either a term the community itself provided, or an explicit rejection.

We make three contributions. First, we describe the MaaL architecture and its core primitive, the \emph{model-typed input}, and argue for its specific relevance to the African-language data problem. Second, we work through a representative structured data-collection scenario -- a closed-vocabulary agricultural survey -- to make the mechanism concrete, and report analytical feasibility estimates for the resulting system. Third, we propose mechanically transpiling the closed-vocabulary elements already present in existing digital form tools (ODK, KoboToolbox) into MaaL schemas, giving low-literacy respondents a path to complete forms unassisted, in their own language, without requiring organisations to build new data-collection infrastructure from scratch.

\section{Related Work}

\textbf{Model as a Service (MaaS)}, the dominant paradigm for accessing large models via cloud APIs, is structurally unsuited to intermittent-connectivity deployment~\cite{maas}. \textbf{Community-trained small language models} are the closest precedent: InkubaLM, a 400M-parameter model trained from curated data across five African languages, matches or exceeds much larger models on in-language tasks~\cite{inkuba}, validating that deliberately sourced small models beat scraped large ones for underrepresented languages. \textbf{Edge AI platforms} such as Edge Impulse package multiple models into a deployable firmware image~\cite{edgeimpulse}, but require cloud retraining and full reflash for every vocabulary change; MaaL's on-device enrolment removes this dependency. \textbf{Mobile data collection platforms} (ODK, CommCare) solve offline form synchronisation but leave AI inference cloud-side~\cite{odk}; MaaL is the missing offline inference layer for exactly these platforms. \textbf{Few-shot and query-by-example keyword spotting} provides the closed-set matching mechanism MaaL depends on: prototypical-network approaches enrol new keywords from a handful of examples without retraining~\cite{parnami2022few}, open-set variants explicitly separate enrolled terms from out-of-vocabulary input~\cite{rusci2023fewshot}, and on-device domain adaptation has been demonstrated under sub-10KB RAM budgets~\cite{cioflan2024ondevice} -- MaaL packages this class of method as a versioned software dependency rather than proposing a new one.

\section{The MaaL Architecture}

A MaaL domain model is a tuple $M = (B, P, L, \tau, m)$, where $B$ is a frozen embedding backbone mapping input of modality $m$ to a fixed-dimensional vector space; $P$ is a prototype store of enrolled keyword embeddings; $L$ is a label map from keywords to normalised output values; and $\tau$ is a confidence threshold below which the model returns an explicit rejection rather than a guess. Inference is nearest-prototype matching: embed the input, compare by cosine similarity against every enrolled prototype, accept if the best match clears $\tau$. This is a closed-set operation by construction -- the model can only return something a community member actually said, or nothing at all. This closed-set property eliminates generative fabrication -- MaaL cannot invent a term no one said -- but it does not eliminate misrecognition: a wrong enrolled term can clear $\tau$ (false accept), and a correctly-spoken term can fall short of it (false reject). These are classification errors, not hallucinations, but in health and agricultural data collection they carry real harm and are not yet analysed here.

The system has three layers (Table~\ref{tab:arch}). A \emph{form schema} declares, for each field, which domain model applies -- the only artefact a developer writes. A \emph{nomadic runtime} plays the prompt, records the response, routes it to the declared model, and applies the threshold, entirely on-device. The \emph{model library} holds one shared embedding backbone (trained once, centrally, on general multilingual speech data) plus many small, independently versioned domain models, each contributing only a prototype store of roughly 512 bytes per enrolled term.

\begin{table}[h]
\centering
\caption{The three-layer MaaL architecture.}
\label{tab:arch}
\begin{tabular}{lll}
\toprule
Layer & Component & Role \\
\midrule
1 & Form schema & Field $\to$ domain-model bindings; no ML code \\
2 & Nomadic runtime & Prompt, record, route, threshold -- fully offline \\
3 & Model library & Shared backbone + versioned, community-enrolled models \\
\bottomrule
\end{tabular}
\end{table}

Enrolling a new term requires a small number of example recordings passed once through the frozen backbone; the new prototype is their mean embedding. No gradient computation, GPU, or connectivity is required, so a field worker can add a locally-specific term -- a crop variety, a market name -- in minutes, directly from the language as spoken by the people who will use the system.

\section{Illustration: From a Text Form to a Voice Form}

To make the mechanism concrete, consider a representative closed-vocabulary structured survey of the kind widely used in agricultural extension and monitoring programmes, with questions such as: which crop was planted, how many units of seed were used, in which season planting occurred, and whether a given input was applied. Each question maps onto one of a small number of reusable domain-model types: \texttt{agri\_term} (a bounded set of crop names, mapped to normalised identifiers), \texttt{numeric} (spoken quantities), \texttt{time\_period} (seasons or months), and \texttt{yes\_no} (affirmatives and negatives). \texttt{numeric} is architecturally distinct from the other three types: spoken quantities are compositional rather than a small closed set, and nearest-prototype matching does not by itself explain how arbitrary multi-digit numbers are handled. A practical approach -- bounding the field to a discretised range of enrolled values, or composing digit-level matches -- is left unspecified here and is a concrete design question, not yet a solved one.
These four types cover a large share of the fields in a typical structured survey and, once built, are reusable across many forms and deployment languages. 

For any given deployment language, the vocabulary for each model -- the specific crop names, season terms, and affirmative/negative pairs actually spoken -- is compiled from community-verifiable sources and confirmed with speakers directly, rather than scraped from web text: the same design choice that makes MaaL resistant to the dialect-flattening failure mode described in Section~1. A term that cannot be confirmed against a reliable source should be left marked for verification rather than guessed, a discipline a generative model has no equivalent mechanism to enforce. At runtime, a spoken response to each question is embedded and matched against the enrolled vocabulary for that field's domain model; an accepted match writes a normalised value into the corresponding form field, and the completed record is assembled and submitted exactly as a text-form submission would be, without any field ever having been read or typed.

\subsection{From Text Forms to Audio-First Forms}

A practical obstacle to adopting any new data-collection paradigm is that organisations across the Global South already have large investments in existing digital form tools -- ODK, KoboToolbox, and similar XForm-based platforms~\cite{odk} are in active use across health, agriculture, and humanitarian programmes, but their interfaces assume the respondent, or an intermediary, can read.

Closed-vocabulary form elements -- an XForm \texttt{<select1>}, a dropdown, an HTML \texttt{<select>} -- already enumerate exactly the bounded vocabulary a MaaL domain model needs: the option list \emph{is} a prototype-store vocabulary, and the declared answer values \emph{are} a label map $L$. A numeric field binds directly to a \texttt{numeric}-type domain model; a yes/no radio group binds to \texttt{yes\_no}. A large class of existing text forms can therefore be \emph{mechanically transpiled} into a MaaL form schema (Table~\ref{tab:arch}, Layer~1): parse the existing form definition, map each closed-vocabulary field to a domain model seeded with its option list, and enrol prototypes for that vocabulary from community recordings, as illustrated above. The organisation's existing form logic and question ordering require no change.

This reframes MaaL as an accessibility layer underneath data-collection infrastructure already deployed across the region, rather than a replacement for it -- turning a form previously completable only with a literate intermediary reading questions aloud into one a respondent can complete unassisted, in their own language, offline. Building and evaluating such a transpiler is a concrete near-term extension of this work.

\section{Feasibility}

We report analytical estimates grounded in published DS-CNN benchmarks, not measurements from a deployed system (Table~\ref{tab:feas}). These estimates cover on-device storage and latency only; they say nothing about recognition accuracy, which depends on few-shot enrolment quality under real acoustic conditions and is not addressed by parameter counts or benchmark footprints. Implementing and evaluating the real backbone on field-collected audio in a target deployment language is the paper's central open question. The DS-CNN family~\cite{hello_edge} spans roughly 39K (DS-CNN-S) to 189K (DS-CNN-M) parameters; after INT8 quantisation, DS-CNN-S measures 52.5\,KB in the MLPerf Tiny reference benchmark~\cite{mlperf_tiny} and 46.15\,KB in an independent hardware measurement on a Raspberry Pi Pico~2~\cite{polish_kws} -- two sources converging on $\sim$46--52\,KB for the smaller variant. Scaling this ratio to DS-CNN-M gives an approximate upper bound of $\sim$190\,KB, an extrapolation rather than a third direct measurement. The resulting footprint and latency figures (Table~\ref{tab:feas}) are comfortably within the storage and compute budget of commodity Android devices and microcontroller-class hardware, and compatible with the natural pacing of a spoken interaction.

\begin{table}[h]
\centering
\small
\caption{Analytical feasibility estimates (literature-grounded, not measured).}
\label{tab:feas}
\begin{tabular}{@{}p{2.6cm}p{1.6cm}p{7.7cm}@{}}
\toprule
Quantity & Estimate & Basis \\
\midrule
On-device footprint & 46--202 KB & DS-CNN-S/M~\cite{hello_edge}, quantised sizes~\cite{mlperf_tiny,polish_kws}, + 4 domain models, 25 terms \\
Latency (Android) & 20--60 ms & DS-CNN operation counts~\cite{hello_edge} \\
Latency (ESP32-S3) & $\approx$160 ms & Edge Impulse KWS benchmark~\cite{edgeimpulse} \\
Update bandwidth & 0.5--50 KB & Delta OTA to full sneakernet/mesh transfer \\
\bottomrule
\end{tabular}
\end{table}

\section{Discussion and Open Questions}

MaaL's narrowness is deliberate: it cannot answer an open question or handle a request outside its enrolled vocabulary, and by design it should not attempt to. Where richer capability is genuinely needed, we favour an explicit, visually distinct escalation to a community-trained generative model such as InkubaLM~\cite{inkuba}, rather than blurring the boundary that gives MaaL its zero-hallucination property. Open questions include: accuracy of enrolled models under real field acoustic conditions rather than synthetic audio, including the false-accept/false-reject tradeoff as a function of $\tau$ and its associated misrecognition harms; the minimum viable shared backbone across a cluster of related regional languages sharing a deployment area; robustness of a text-to-audio-form transpiler (Section~4.1) across heterogeneous existing schemas; and, most centrally for this venue, what governance structure should oversee community vocabulary curation itself -- who decides what a term means, who may extend it, and how disagreement about a translation or dialect boundary is resolved, rather than settled implicitly by whoever built the tool.

\section{Conclusion}

Large language models are least reliable exactly where linguistic need is greatest, and the cause is a data gap that scale alone cannot fix. MaaL proposes a narrower but more honest alternative: small, community-enrolled, closed-set models that cannot misrepresent what was not said, work fully offline, and grow their vocabulary from the language as it is actually spoken rather than from a scraped corpus. Treating the closed-vocabulary structure already present in widely-deployed digital form tools as a ready-made source of that vocabulary also offers a route to adoption that does not require organisations to discard existing infrastructure.

\begin{ack}

This work received no dedicated research grant funding from any agency in the public, commercial, or not-for-profit sectors. The author gratefully acknowledges the Stellenbosch Institute for Advanced Study (STIAS), Stellenbosch, South Africa, for the fellowship and residency during which this work was developed.

We used Claude (Anthropic) to assist with drafting and editing this manuscript, deriving the analytical estimates in Table~\ref{tab:feas} from cited literature values, and researching and verifying citations. All citations, technical claims, and derivations were independently checked against primary sources before submission. The MaaL architecture, the model-typed input primitive, and the audio-first-forms proposal originate from the author; no LLM is a component of the proposed system.

\end{ack}

\printbibliography

\end{document}